\documentclass[10pt,twocolumn,letterpaper]{article}

\pdfoutput=1
\usepackage[pagenumbers]{cvpr} % To force page numbers, e.g. for an arXiv version

\usepackage{xspace}
\newcommand{\modelname}{LRDet}   % <-- change only this
\newcommand{\modelT}{\modelname-T\xspace}          % prose
\newcommand{\modelB}{\modelname-B\xspace}          % prose
\newcommand{\oursT}{\modelname-T (ours)}           % tables
\newcommand{\oursB}{\modelname-B (ours)}           % tables

\usepackage{cuted}
\usepackage{tikz}
\usepackage{amsmath}
\usetikzlibrary{arrows.meta}
\usepackage{pgfplots}
\usepgfplotslibrary{groupplots}
\usetikzlibrary{calc}

\usepackage{microtype}

\renewcommand{\paragraph}[1]{\vspace{.5em}\noindent\textbf{#1.}}

\definecolor{cvprblue}{rgb}{0.21,0.49,0.74}
\usepackage[pagebackref,breaklinks,colorlinks,allcolors=cvprblue]{hyperref}
\usepackage{needspace}

\def\confName{CVPR}
\def\confYear{2027}

\title{\vspace{-8mm}The Role of Recurrent Memory in Event-Based Vision}

\author{
Xiao Xiang\textsuperscript{1,2,3*}\quad
Johan Lindqvist\textsuperscript{1,2,3}\quad
Tim Brockmeyer\textsuperscript{4 $\dagger$}\quad
Antonio Orvieto\textsuperscript{1,2,3 $\dagger$}\\[2pt]
{\normalsize
\textsuperscript{1}ELLIS Institute T\"ubingen\quad
\textsuperscript{2}Max Planck Institute for Intelligent Systems\quad
\textsuperscript{3}T\"ubingen AI Center\quad
\textsuperscript{4}Airbus}\\[2pt]
}

\begin{document}

\twocolumn[{%
  \renewcommand\twocolumn[1][]{#1}%
  \maketitle
  \vspace{-4.7mm}                      % gap: authors -> teaser
  \centering
  \begin{minipage}[t]{0.65\textwidth}
    \vspace{0pt}\centering
    \resizebox{\linewidth}{!}{\input{tex/teaser}}
  \end{minipage}\hfill
  \begin{minipage}[t]{0.33\textwidth}
    \vspace{0pt}
    \vspace{1.2mm}
    \centering
    \makebox[\linewidth][r]{\resizebox{1.108\linewidth}{!}{\definecolor{memoryteal}{RGB}{0,125,122}
\definecolor{evred}{RGB}{214,39,40}
\definecolor{evblue}{RGB}{31,119,180}
\begin{tikzpicture}
\begin{axis}[
  width=6.65cm, height=8.55cm, clip=false,
  xmin=0.7, xmax=8.3, ymin=0, ymax=30,
  xtick={1,...,8}, ytick={0,5,...,30},
  xlabel={Time since last visible (s)}, ylabel={AP},
  ymajorgrids, grid style={gray!25},
  axis lines=left, tick align=outside,
  tick label style={font=\rmfamily\bfseries\fontsize{11}{12}\selectfont\boldmath},
  label style={font=\rmfamily\bfseries\fontsize{12}{13}\selectfont\boldmath},
  every axis plot/.append style={line width=1.1pt, mark size=1.8pt},
]

\fill[evred, opacity=.10] (axis cs:1,12.78) -- (axis cs:2,4.52) -- (axis cs:3,2.71) -- (axis cs:4,1.57)
  -- (axis cs:5,1.56) -- (axis cs:6,0.78) -- (axis cs:7,0.67) -- (axis cs:8,0.64)
  -- (axis cs:8,6.23) -- (axis cs:7,6.28) -- (axis cs:6,6.47) -- (axis cs:5,8.19) -- (axis cs:4,8.38)
  -- (axis cs:3,10.27) -- (axis cs:2,13.80) -- (axis cs:1,23.06) -- cycle;
\fill[evblue, opacity=.12] (axis cs:1,23.06) -- (axis cs:2,13.80) -- (axis cs:3,10.27) -- (axis cs:4,8.38)
  -- (axis cs:5,8.19) -- (axis cs:6,6.47) -- (axis cs:7,6.28) -- (axis cs:8,6.23)
  -- (axis cs:8,12.15) -- (axis cs:7,12.27) -- (axis cs:6,13.43) -- (axis cs:5,14.87) -- (axis cs:4,15.65)
  -- (axis cs:3,17.20) -- (axis cs:2,20.16) -- (axis cs:1,27.66) -- cycle;
\addplot[evblue, mark=*] coordinates {
 (1,27.66) (2,20.16) (3,17.20) (4,15.65) (5,14.87) (6,13.43) (7,12.27) (8,12.15)};
\addplot[evred, mark=square*] coordinates {
 (1,23.06) (2,13.80) (3,10.27) (4,8.38) (5,8.19) (6,6.47) (7,6.28) (8,6.23)};
\addplot[memoryteal, mark=triangle*] coordinates {
 (1,12.78) (2,4.52) (3,2.71) (4,1.57) (5,1.56) (6,0.78) (7,0.67) (8,0.64)};
\node[anchor=west, evblue, font=\rmfamily\bfseries\fontsize{12}{13}\selectfont] at (axis cs:8.2,12.15) {Ours};
\node[anchor=west, evred,  font=\rmfamily\bfseries\fontsize{12}{13}\selectfont] at (axis cs:8.2,6.23) {Linear attention};
\node[anchor=west, memoryteal, font=\rmfamily\bfseries\fontsize{12}{13}\selectfont] at (axis cs:8.2,0.64) {No memory};

\draw[->, >=stealth, line width=1.4pt] (axis cs:4.5,1.6) -- (axis cs:4.5,8.1);
\draw[->, >=stealth, line width=1.4pt] (axis cs:4.5,8.5) -- (axis cs:4.5,15.0);
\node[align=center, font=\rmfamily\bfseries\fontsize{9.5}{11}\selectfont] (good) at (axis cs:6.4,10.2) {What makes a\\good memory?};
\node[align=center, font=\rmfamily\bfseries\fontsize{9.5}{11}\selectfont] (need) at (axis cs:6.6,3.4) {Do we need\\memory?};
\draw[black!60] (good.west) -- (axis cs:4.55,10.2);
\draw[black!60] (need.west) -- (axis cs:4.55,3.4);
\end{axis}
\end{tikzpicture}}\hspace*{-0.1cm}}
  \end{minipage}
  \captionsetup{hypcap=false}
  
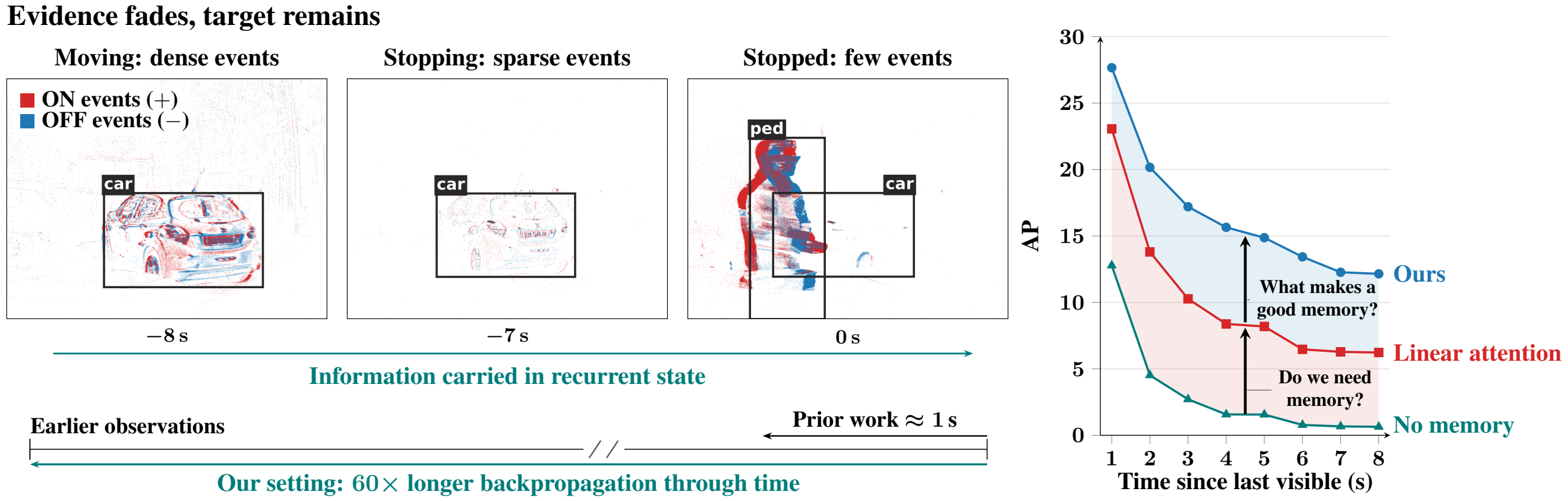
\captionof{figure}{\textbf{A detector must remember what it no longer
  senses.} Left: an event camera records when and where pixel brightness
  increases or decreases. A car stops moving and leaves almost no events, yet
  it remains in the scene. Boxes show the ground truth. Right: AP on Gen1
  validation for objects that are not visible, by the time since they were
  last visible. An object is visible if at least 10\% of its box pixels emit
  an event. Without memory, detection collapses, and our model retains about
  twice as much as linear attention at the longest gaps shown.}
  \label{fig:intro}
  \vspace{3mm}                       % gap: caption -> abstract
}]

\begingroup
\renewcommand\thefootnote{}
\footnotetext{\textsuperscript{*}Correspondence: {\tt xiao.xiang@tuebingen.mpg.de}}
\footnotetext{\textsuperscript{$\dagger$}Equal supervision}
\endgroup

\begin{abstract}
Memory is essential for event-based vision. Event cameras respond to changes in brightness and objects with little motion relative
to the camera can generate few events. Detection thus relies on memory from earlier observations. Prior recurrent detectors address this challenge, but truncate backpropagation through time to approximately one second of events. We lift this limit with a $60\times$ longer gradient horizon and train decoupled temporal modules on frozen spatial features. To process temporal information, we compare recently proposed state-space models and stateful linear attention variants, which differ in their update rules governing what the model can retain and forget over time, as well as in their state-tracking capabilities. We introduce an event-based shell game to inspect permutation tracking performance, which has been linked to improved reasoning capabilities in large language models. We find that models that perform best in this controlled setting are also the strongest detectors on event recordings, particularly when current evidence is insufficient. Stacking these recurrent blocks with depth-scaled training further improves detection, in line with theoretical guarantees established for state tracking. The resulting detector outperforms existing methods on Gen1 and 1Mpx.
\end{abstract}    
\section{Introduction}
\label{sec:intro}
Event cameras take inspiration from biological vision, responding to changes in light intensity \cite{Survey}. Biological perception additionally integrates sensory information over time \cite{cortex} supported in part by recurrent processing \cite{recurrence_brain}. Object permanence, the notion that we expect objects to persist even when they are no longer visible, illustrates a broader role of memory in visual tasks. Earlier observations help us infer what is still present. Event-based detection faces a similar challenge (Figure~\ref{fig:intro}). When the target remains in the scene without moving, it generates few or no events. Detecting such targets therefore depends on information from earlier observations, potentially separated by many event frames even within a brief recording.

Linear recurrences offer a way to carry this sequence of evidence forward through associative memory \cite{correlation_matrix,linear_attention, Fast_weight}. They are more stable to train than nonlinear
RNNs \cite{lru,zucchet}, parallelize over the sequence, and
compete with softmax attention at scale. These architectures have recently seen a resurgence
\cite{lru,mamba2,gdn}, with large language models such as Kimi K3, Qwen4-Max and Olmo Hybrid adopting variants of linear attention to model long contexts efficiently \cite{kimi_k3, kda, qwen4, olmo_hybrid}, and found that they offers expressivity benefits beyond softmax attention \cite{olmo_hybrid}. This success motivates examining how the state transitions of linear recurrent architectures translates to event-based vision.

The expressiveness of the state transition in a linear recurrent layer determines what is \textbf{retained}, \textbf{forgotten}, or \textbf{overwritten}. Theoretical work also links the structure of this transition to state tracking, a capability that supports reasoning tasks in language models, such as following a sequence of moves in a game \cite{illusion_ssm, olmo_hybrid}. The challenge of detection gives us a concrete setting to inspect these capabilities: \textbf{What is the role of recurrent memory in event-based vision?}

This exact recurrent memory also needs to bridge several seconds of weak or absent evidence \cite{rvt}. Prior work, such as Recurrent Vision Transformer \cite{rvt} recognizes that slowly moving objects generates few events, making features from several seconds earlier important for detection, and integrate spatial and temporal processing within its backbone. EvRT-DETR \cite{evrt} further decouples the two, freezing a frame-based detector \cite{detr} and adding a lightweight ConvLSTM \cite{convlstm} on its encoder features to carry information across time. In both, the recurrent state carries over through long recordings. However, backpropagation is truncated at approximately one second, substantially below the temporal horizon required by many slowly moving objects. To lift this limit, we make the following contributions:
\begin{itemize}
    \item We propose a setup for studying recurrent memory in event-based
    detection, leveraging the parallelization properties of linear recurrent models to use a $60\times$ longer gradient horizon than prior work \cite{rvt,evrt}.
    (Sec.~\ref{sec:method}). The resulting detector
    outperforms existing methods on Gen1 \cite{Gen1} and 1Mpx \cite{1mpx} (Tab.~\ref{tab:main}).

\item Guided by the theoretical properties of linear recurrences, such as state tracking, we study variants from basic linear attention to KDA \cite{kda,CKDA} (Sec.~\ref{sec:Uniformized_backbone}), and find that detection performance is improved by expressiveness, most clearly when current evidence is insufficient (Tab.~\ref{tab:ablation}).

\item We revisit how temporal modules allocate additional parameters, and find that depth
scales more efficiently for event-based detection than larger spatial kernels. We discuss how stacking recurrent layers increases expressivity (Sec.~\ref{sec:depth_method})  and identify learning rate settings that make deeper stacks trainable (Sec.~\ref{sec:depth}).

\item We probe state tracking with a toy event-based shell game, in
which models follow object identities through swaps after identity cues
fade. Only recurrences with delta updates show signs of tracking, and depth extends this ability to longer swaps than seen
in training (Sec.~\ref{sec:shell_game}).

\end{itemize}

\section{Related work}
\label{sec:related work}

\subsection{Recurrent Model for Event-Based Detection}
Object detection is the task of predicting a bounding box and a class
label for each object of interest in a scene. It is among the most studied problems in computer vision, and represents some of the first challenges of scene understanding \cite{coco}. Over the past decade, detectors have evolved from region proposal \cite{rcnn,faster_rcnn} through single-shot detectors \cite{yolo,yolox} to transformer-based set prediction \cite{detr}.

Event-based detection has followed a similar path, with benchmarks such as Gen1 \cite{Gen1}, 1Mpx \cite{1mpx}, and, more recently, eTraM \cite{etram}. RED \cite{1mpx} places ConvLSTM layers \cite{convlstm} after a convolutional backbone to integrate information across event windows. Recurrent Vision Transformer (RVT) further pairs a MaxViT-style spatial backbone \cite{maxvit} with LSTM layers that share weights across spatial positions and maintain a separate state at each position \cite{rvt}. This line of work has recently adopted DETR-style detector \cite{detr}, with EvRT-DETR adding ConvLSTM modules to a frozen frame detector \cite{evrt}.

SSM-ViT \cite{SSM-ViT} builds on RVT by replacing its LSTM layers with linear state-space layers \cite{s4d,s5}. At inference, the same model runs on event windows of a different
duration by rescaling its discretization step, without retraining
\cite{SSM-ViT}.

\subsection{Linear Recurrent Models}
\label{rel}
The state space models used in SSM-ViT \cite{SSM-ViT} belong to a broader family of linear recurrent models. As their memory updates are affine in the previous state, they support parallel computation across time steps during training and recurrent inference with cost linear in sequence length~\cite{s5,mamba}. For a fixed sequence length, a sufficiently wide diagonal recurrence
with a nonlinear readout can approximate a broad class of causal
sequence-to-sequence functions to arbitrary accuracy~\cite{universality}. This combination of efficiency and expressivity was already illustrated in early structured SSMs, which outperformed LSTM variants across a range of long-sequence benchmarks~\cite{lssl,s4}.

Structured transitions also allow the recurrent state to grow at a lower parameter cost than widening a conventional LSTM. This increases the capacity available to retain past observations within a fixed parameter budget~\cite{mamba,xlstm}. Depth offers a complementary way to scale linear recurrent models. In a stack of recurrent layers, each layer processes features that already incorporate information from earlier time steps~\cite{lru,mamba}. For selective SSMs, this composition broadens the class of functions of the input history that the model can represent ~\cite{circone}.

Recent linear recurrent models offer different ways to retain and
revise information over time~\cite{s5,mamba,mamba2,gdn,kda}. We discuss how these recurrent updates differ and compare their
effects on detection using a shared temporal block and identical
spatial features (\cref{sec:Uniformized_backbone}).

\begin{figure*}[t]
  \centering
  \input{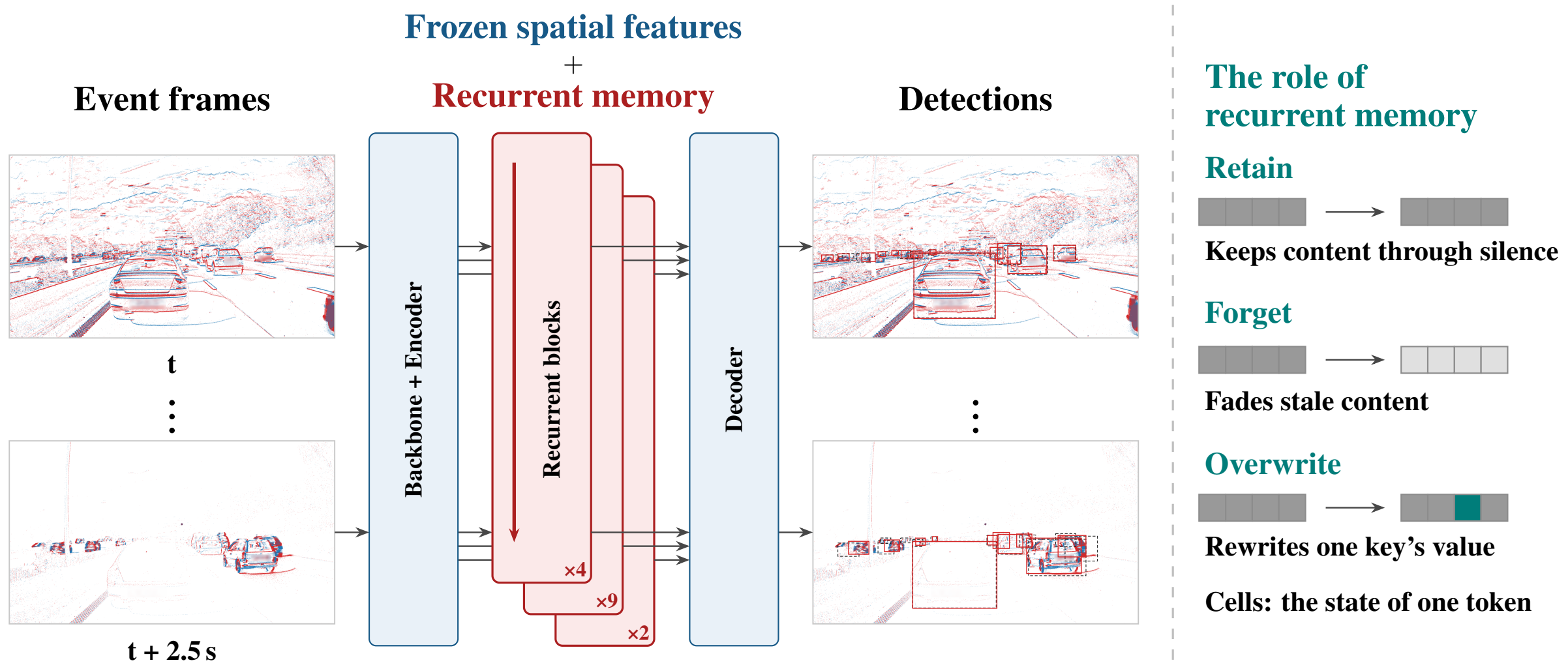}
 \caption{\textbf{A recurrent memory on frozen spatial features.}
Left: the \textcolor{evblue}{frozen backbone, encoder and decoder} of a frame
detector, with a \textcolor{evred}{recurrent memory} of 4, 9 and 2
blocks at strides 8, 16 and 32. Ground truth is dashed and our detections are in red, with class labels omitted. Right: we illustrate how the state of one spatially mixed token is
\textcolor{memoryteal}{retained, forgotten or overwritten}
(Sec.~\ref{sec:Uniformized_backbone}).}
  \label{fig:pipeline}
\end{figure*}

\subsection{State Tracking in Vision}

State tracking is the task of determining the current state of a system from a sequence of updates, each acting on the state left by the previous one~\cite{illusion_ssm}. In language, this involves following changes to an entity's location or properties \cite{kim2023entity}, a capability argued to contribute to the strong performance of recurrent models for language modeling \cite{olmo_hybrid}. Transition structure decides which changes a model can track. Under standard complexity assumptions, fixed-depth diagonal SSMs
with logarithmic precision cannot solve certain permutation-tracking
tasks over arbitrary sequence lengths~\cite{illusion_ssm}. Extensions
to the delta rule broaden the range of state-tracking tasks that
linear recurrent models can solve~\cite{negative_eigen,CKDA}.

Visual object permanence illustrates why state tracking matters. A hidden object’s last observed location remains valid while its container is stationary, but its location must be inferred from the container’s motion once it moves \cite{shamsian2020learning}. The choice of recurrent transition can affect how reliably a model follows such hidden changes \cite{shin2026video}. Recent benchmarks pose this problem as a shell game, in which a hidden
object has to be followed through swaps of identical containers
\cite{vet_bench,cohen2026metbenchmultimodalentitytracking}. In event-based vision, both occlusion and low relative motion can interrupt direct evidence of an object \cite{1mpx}. As motion resumes, earlier observations may still indicate what is present while new events along its boundary indicate where it has moved. This motivates studying how recurrent detectors preserve object information across gaps and revise it as new evidence arrives.

\section{Method}
\label{sec:method}
In this section, we illustrate our design trajectory from a conventional frame detector to a detector with a well-designed recurrent memory. We first introduce a setup that extends the gradient horizon and
enables controlled comparisons of recurrent memories over long
recordings (\cref{sec:framework}). We then present a uniformized temporal backbone in which the memory mechanisms differ only in their state transitions (\cref{sec:Uniformized_backbone}). In \cref{sec:depth_method}, we return to a commonly tested axis in temporal module design in event-based vision, the spatial kernel of the memory, and explain our choice to scale the depth of a recurrent memory.

\subsection{A Detector with Long-Horizon Memory}
\label{sec:framework}
\paragraph{Events to stacked histograms}
The input is a recording from an event camera, an asynchronous stream of
per-pixel brightness changes. We partition the stream into windows of
50\,ms and accumulate each window into a stacked histogram $x_t \in \mathbb{R}^{20 \times H \times W}$ of the events
over across two polarities and ten time bins, following \citet{rvt}. This turns the event stream
into a sequence of dense tensors that an image detector can process. A recording is
thus a sequence $x_1, \dots, x_T$. For example, a recording of 60\,s,
the length used in Gen1 \cite{Gen1} and 1Mpx \cite{1mpx}, gives
$T = 1200$.

\paragraph{Event frame detector}
We start from RT-DETR \cite{detr}, a detector designed for images and
trained on individual event frames \cite{evrt}. Its backbone extracts features at four scales \cite{resnet}, the last three of which
are fused by the encoder. Together, the two constitute the spatial
module $f$, which maps each frame to multi-scale features,
\begin{equation*}
  \{E_t^{\ell}\}_{\ell=3}^{5} = f(x_t), \qquad
  E_t^{\ell} \in \mathbb{R}^{C \times H_\ell \times W_\ell},
\end{equation*}
with $C$ channels and strides of 8, 16 and 32. Its transformer decoder is
the detection head, which predicts a set of bounding boxes and their
classes. In our work, we freeze the spatial module $f$ and the decoder.

% In
% the second stage, $f$ and the decoder are frozen, and a temporal module
% is bridged between the two components. The decoder then reads features $\tilde E_t^{\ell}$ that carry information from all frames up to $t$.

\paragraph{Temporal Module} 
Progressing from a frozen detector, we place a temporal module between
$f$ and the decoder similar to \citet{evrt}. This turns each $E_t^{\ell}$ into a feature map
$\tilde E_t^{\ell}$ that carries recurrent memory from all frames up to $t$,
and the decoder reads $\tilde E_t^{\ell}$ in place of $E_t^{\ell}$. 

At each scale, every spatially mixed token at one location $u_t = E_t^{\ell}[p] \in \mathbb{R}^{C}$ forms the input sequence of a linear recurrent model. One model per scale is shared across all locations of the feature map.

\paragraph{Gradient horizon} 
Truncating backpropagation through time limits how far back the memory is trained to look. Prior recurrent detectors \cite{rvt,SSM-ViT,evrt} carry the state across recording, with backpropagation limited to 21 frames
on Gen1~\cite{Gen1} and 5--10 frames on 1Mpx~\cite{1mpx}. This horizon of at most one second is bounded by two costs. A 3×3 ConvLSTM processes frames sequentially~\cite{convlstm}, whereas its linear recurrent counterpart computes updates in parallel across time, yielding a 3 to 17x speedup in the forward pass over a long recording, depending on the configuration and task. Joint spatial-temporal training in RVT \cite{rvt,SSM-ViT} additionally stores backbone activations per frame, so its training memory grows with the clip length at the cost of the whole backbone. Our design avoids both costs, with linear
recurrence architectures trained in parallel over time and a frozen spatial weights. We train on whole 60\,s recordings on Gen1 and 30\,s clips on 1Mpx, $60\times$ the prior horizon.

\subsection{Recurrent Update in a Uniformized Backbone}
\label{sec:Uniformized_backbone}

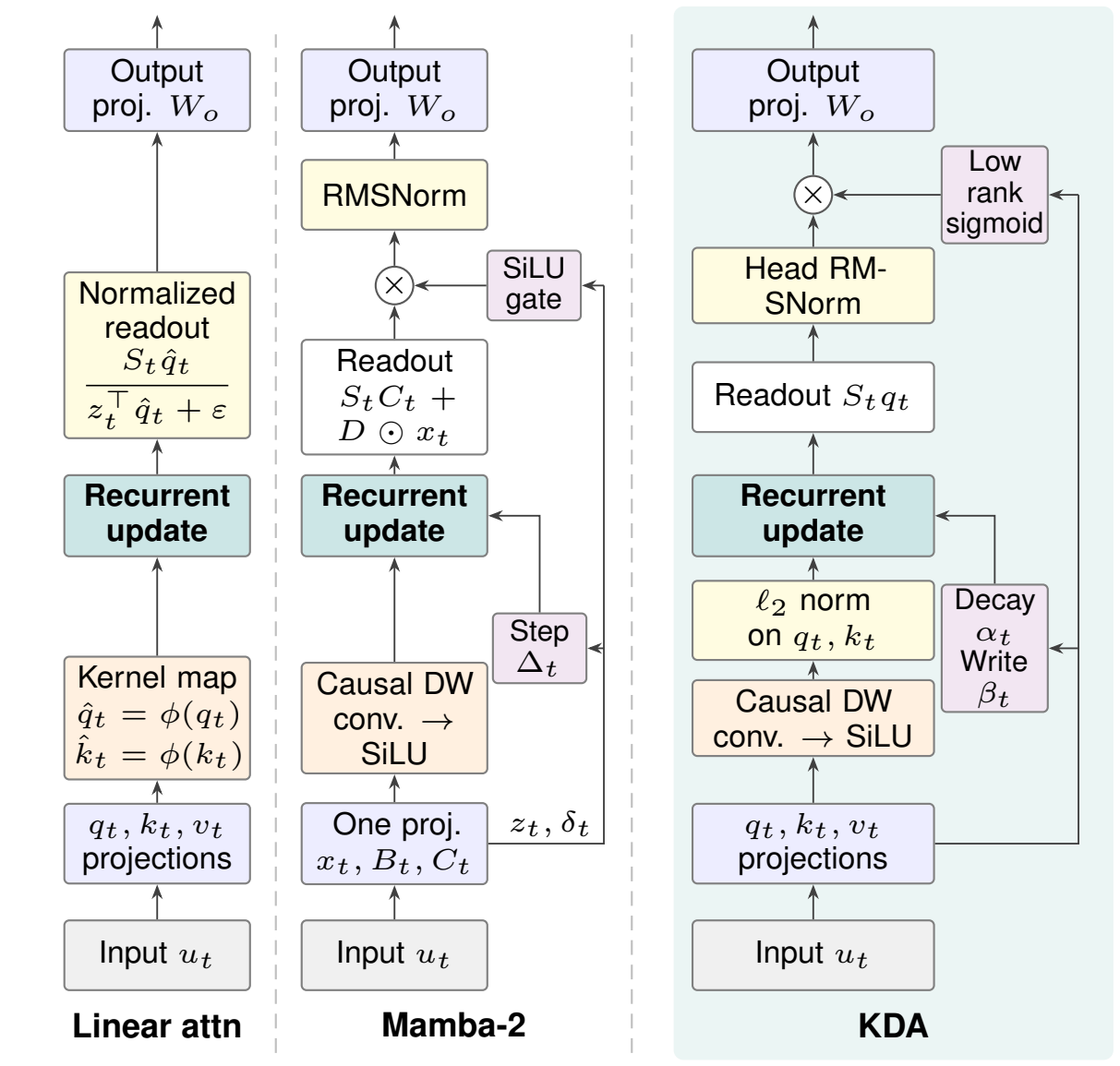
\begin{figure}[t]
  \centering
  \providecolor{memoryteal}{RGB}{0,125,122}
\begin{tikzpicture}[
  x=1mm,y=1mm,font=\sffamily\fontsize{6.2}{7.1}\selectfont,
  >={Stealth[length=1.05mm]},
  flow/.style={->,draw=black!75,line width=.35pt},
  stage/.style={draw=black!50,rounded corners=1pt,align=center,
    text width=15mm,minimum height=5mm,inner sep=.65mm,fill=white},
  proj/.style={stage,fill=blue!7},
  prep/.style={stage,fill=orange!12},
  state/.style={stage,fill=teal!20},
  gate/.style={stage,fill=violet!10,text width=6.1mm,
    inner sep=.3mm,font=\sffamily\fontsize{5.8}{6.3}\selectfont},
  norm/.style={stage,fill=yellow!14},
  op/.style={circle,draw=black!65,fill=white,
    inner sep=0pt,minimum size=2.7mm}
]

\fill[memoryteal!8,rounded corners=2pt] (53,-7.5) rectangle (84.5,68);

\foreach \i/\model/\sh/\cx/\w/\tx in {0/Linear attn/7/9/12/9, 1/Mamba-2/24/9/12/13.25, 2/KDA/53/10/16/15.75}{
  \begin{scope}[xshift=\sh mm]
        \node[font=\sffamily\bfseries\fontsize{6.8}{7.5}\selectfont] at (\tx,-5) {\model};
    \node[stage,fill=black!6,text width=\w mm] (in\i) at (\cx,0) {Input $u_t$};
    \ifnum\i=1
      \node[proj,text width=\w mm] (pr\i) at (\cx,8) {One proj.\\$x_t,B_t,C_t$};
    \else
      \node[proj,text width=\w mm] (pr\i) at (\cx,8) {$q_t,k_t,v_t$\\projections};
    \fi
    \node[state,text width=\w mm] (mix\i) at (\cx,31.5) {\textbf{Recurrent update}};
    \node[proj,text width=\w mm] (out\i) at (\cx,62) {Output proj. $W_o$};
    \draw[flow] (in\i)--(pr\i);
    \draw[flow] (out\i.north)--++(0,2.4);
  \end{scope}
}
\draw[black!25,densely dashed] (24.5,-7)--(24.5,66);
\draw[black!25,densely dashed] (50,-7)--(50,66);

\begin{scope}[xshift=7mm]
  \node[prep,text width=12mm] (lf) at (9,17)
    {Kernel map\\$\hat q_t=\phi(q_t)$\\$\hat k_t=\phi(k_t)$};
  \node[norm,text width=12mm] (lr) at (9,43) {Normalized\\readout\\[1pt]
    $\dfrac{S_t\hat q_t}{z_t^\top\hat q_t+\varepsilon}$};
\end{scope}
\foreach \a/\b in {pr0/lf,lf/mix0,mix0/lr,lr/out0} \draw[flow] (\a)--(\b);

\begin{scope}[xshift=24mm]
  \node[prep,text width=12mm] (mc) at (9,17) {Causal DW\\conv. $\to$ SiLU};
  \node[gate] (md) at (19.4,22) {Step\\$\Delta_t$};
  \node[stage,text width=12mm] (mr) at (9,40) {Readout\\$S_t C_t+D\odot x_t$};
  \node[op] (mm) at (9,48) {$\times$};
  \node[gate] (mg) at (19,48) {SiLU\\gate};
  \node[norm,text width=12mm] (mn) at (9,54.5) {RMSNorm};
  \foreach \a/\b in {pr1/mc,mc/mix1,mix1/mr,mr/mm,mm/mn,mn/out1,mg/mm}
    \draw[flow] (\a)--(\b);
  \draw[black!75,line width=.35pt] (pr1.east)--(24,8)--(24,48);
  \node[font=\sffamily\fontsize{5.8}{6.3}\selectfont,above,inner sep=.4mm]
    at (20.2,8) {$z_t,\delta_t$};
  \draw[flow] (24,22)--(md.east);
  \draw[flow] (24,48)--(mg.east);
  \draw[flow] (md.north)|-(mix1.east);
\end{scope}

\begin{scope}[xshift=53mm]
  \node[prep,text width=16mm] (kc) at (10,17) {Causal DW conv. $\to$ SiLU};
  \node[norm,text width=16mm] (kl) at (10,24) {$\ell_2$ norm on $q_t,k_t$};
  \node[gate,text width=7mm] (kd) at (23,22) {Decay $\alpha_t$\\Write $\beta_t$};
  \node[stage,text width=16mm] (kr) at (10,40) {Readout $S_t q_t$};
  \node[norm,text width=16mm] (kn) at (10,48) {Head RMSNorm};
  \node[op] (km) at (10,54.5) {$\times$};
  \node[gate,text width=7mm] (kg) at (23,54.5) {Low rank\\sigmoid};
  \foreach \a/\b in {pr2/kc,kc/kl,kl/mix2,mix2/kr,kr/kn,kn/km,km/out2,kg/km}
    \draw[flow] (\a)--(\b);
  \draw[black!75,line width=.35pt] (pr2.east)--(29,8)--(29,54.5);
  \draw[flow] (29,22)--(kd.east);
  \draw[flow] (29,54.5)--(kg.east);
  \draw[flow] (kd.north)|-(mix2.east);
\end{scope}

\end{tikzpicture}
  \caption{\textbf{Published blocks differ in more than the recurrent
update.} The blocks of linear attention, Mamba-2 and KDA process
features differently before and after the update. We use the KDA block
(shaded) for every recurrence we compare.}
  \label{fig:mixer_architectures}
\end{figure}

To study the role of memory in event-based detection, we introduce a shared
block (\cref{fig:mixer_architectures}) and vary only the recurrent
update. The theory of linear recurrences characterizes what each form
of update allows a memory to do, from retaining and forgetting to
overwriting and tracking a changing state
\cite{illusion_ssm, negative_eigen}. We therefore build the update up step by step guided by these properties, and later explore whether and how detection benefits empirically from each step in \cref{sec:ablation}.

\paragraph{Shared block structure}
Linear recurrent architectures differ both in how they update memory
and in how they process features before and after each update, and these surrounding layers have
been reported to affect performance substantially \cite{physics_lm_4_1,gdn}. Comparing them as published
would therefore confound the update with the block around it. To isolate the effect of recurrent updates, we use
the block of Kimi Delta Attention \cite{kda} for every recurrence we
compare (\cref{fig:mixer_architectures}). Each query, key, and value projection passes through a short causal convolution to incorporate recent observations, followed by a SiLU activation \cite{mamba,deltanet}. We normalize the resulting queries and keys before they enter the mixer. After the update, the query reads from memory, and the readout passes through head-wise RMS normalization, an input-dependent gate, and an output projection. Consistent with common language-model practice, we pair each recurrent mixer with a SwiGLU MLP with residual connections around each unit. %and stack these blocks to build the temporal model.

\paragraph{Linear attention} To understand how the linear recurrent models retain and revise past information, we express their memory updates in a common notation and incorporate one mechanism at a time. We consider a single head with queries and keys \(q_t,k_t\in\mathbb R^{d_k}\), values \(v_t\in\mathbb R^{d_v}\), and a memory state \(S_t\in\mathbb R^{d_v\times d_k}\). Omitting the output normalizer, causal linear attention takes the form
\begin{equation}
S_t=S_{t-1}+v_tk_t^\top
\tag{\textbf{Base}}
\end{equation}
Each outer product writes a rank-one association and accumulates information into the state. At readout, $o_t = S_t q_t$, a past input contributes equally regardless of distance.

\paragraph{Input-independent decay} A first extension gives each head a memory timescale by multiplying its previous state by a constant retention factor,
\begin{equation}
  S_t = \alpha\, S_{t-1} + v_t k_t^\top,
  \tag{\textbf{+ Decay}}
\end{equation}
where $\alpha \in (0, 1)$ controls how fast the memory fades.
A write from step \(i\) is scaled by \(\alpha^{t-i}\), so its retention depends only on its age. In the literature, this includes S5, which learns a complex eigenvalue per state dimension \cite{s5}, and
RetNet, which fixes one real decay per head \cite{retnet}. Our implementation learns one
real decay per head, which we refer to as LTI (Linear time-invariant) Mamba-2.

\paragraph{Selectivity} State space models with fixed dynamics lagged behind
attention on language tasks, a weakness that Mamba attributes to their
inability to adapt to input \cite{mamba}. Mamba makes retention selective by letting the current input determine the step size \(\Delta_t\) \cite{mamba,mamba2}. In our notation,
\begin{equation}
S_t=\alpha_t S_{t-1}
    +v_tk_t^\top,
\tag{\textbf{+ Selectivity}}
\end{equation}
where \(\alpha_t=\exp(-A\Delta_t)\) and
\(\Delta_t=\operatorname{softplus}(w_\Delta^\top u_t+b_\Delta)\).
With learned rate \(A>0\), a small \(\Delta_t\) preserves most of the
previous state, while a larger \(\Delta_t\) increases forgetting.

\paragraph{Delta rule}
Gated DeltaNet (GDN) and Kimi Delta Attention (KDA) combine selective
decay with a precise overwriting operation provided by the delta rule
\cite{Fast_weight,deltanet,gdn,kda}, applied as
\begin{equation}
S_t= \boldsymbol{\alpha_t} S_{t-1}
    \underbrace{(I-\beta_t k_tk_t^\top)}_{\text{erase along $k_t$}}
    +\beta_t v_tk_t^\top,
\tag{\textbf{+ Delta rule}}
\end{equation}
where $\alpha_t$ denotes the selective decay rate, applied per head in GDN and per channel in KDA, and \(\beta_t\) is a writing strength predicted from the input. Small changes to these rules provide significant benefits for expressivity. Extending the correction range of the delta rule
to \(\beta_t\in(0,2)\) permits negative transition eigenvalues \cite{negative_eigen} while additionally increasing the range of \(\alpha_t\) to \((-1,1)\) enables complex eigenvalues \cite{CKDA}. These changes broaden the class of state-tracking tasks it can solve and together expand the expressivity of a single layer to solve certain \(\mathsf{NC}^{1}\)-complete state-tracking problems \cite{negative_eigen,CKDA}.

\subsection{Scaling the Depth of Recurrent Memory}
\label{sec:depth_method}
Prior work has explored larger convolutional kernels to expand the spatial receptive field of the recurrent update 
\cite{convlstm, 1mpx,rvt, get, evrt, smamba}. The reported effect of kernel size varies across detectors. For example, the
pointwise LSTM performs better in RVT \cite{rvt}, while the
$3\times3$ ConvLSTM yields a small gain in EvRT-DETR \cite{evrt}. We revisit this choice on frozen spatial features and propose scaling the linear recurrence in depth instead. Deeper models are known to be more expressive as each further block updates its state
from features that already carry information from earlier frames, and
stacked recurrences can track states that a single layer cannot
\cite{negative_eigen,deltaproduct}. We evaluate both design choices in \cref{sec:depth}, and we expand our discussion on its benefits in \cref{sec:shell_game}.

\section{Experiments}
\label{sec:experiments}
\begin{table*}[t]
\centering
\setlength{\tabcolsep}{8pt}
\begin{tabular}{llccccc}
\toprule
& & \multicolumn{2}{c}{Gen1} & \multicolumn{2}{c}{1Mpx} & Params \\
\cmidrule(lr){3-4}\cmidrule(lr){5-6}
Model & Memory & mAP (\%) & Runtime (ms) & mAP (\%) & Runtime (ms) & (M) \\
\midrule
\multicolumn{7}{l}{\textbf{No memory}} \\
ERGO-12 \cite{ergo}     & --       &50.4 &(69.9) &40.6 &(100.0) &59.6 \\
RT-DETR-T \cite{detr}  & --       &46.0 &4.5  &44.1 &4.7   &20.1 \\
RT-DETR-B \cite{detr}  & --       &47.6 &6.7 &45.2 &6.9  &42.8 \\
\midrule
\multicolumn{7}{l}{\textbf{Memory in the backbone}} \\
RED \cite{1mpx}        & ConvLSTM &40.0 &(16.7) &43.0 &(39.3)        &24.1 \\
ASTMNet  \cite{astmnet}   & Rec-Conv &46.7 &(35.6) &48.3 &(72.3)        &$>$100 \\
RVT-B  \cite{rvt}     & LSTM     &47.2 &4.8 &47.4 &4.8        &18.5 \\
GET-T   \cite{get}    & ConvLSTM &47.9 &(16.8) &48.4 &(18.2)        &21.9 \\
S5-ViT-B  \cite{SSM-ViT}  & S5       &47.4 &7.0$^\dagger$  &47.2 &7.0$^\dagger$         &17.5 \\
SMamba \cite{smamba}     & ConvLSTM      &50.4 &23.7$^\ddagger$ &49.3 &30.5$^\ddagger$        &16.7 \\
SATE    \cite{sate}    & ConvLSTM &52.7 &5.4$^{\dagger\ddagger}$ &49.1 &5.4$^{\dagger\ddagger}$ &26.4 \\
\midrule
\multicolumn{7}{l}{\textbf{Memory on frozen spatial weights}} \\
EvRT-DETR-T \cite{evrt}& ConvLSTM &52.3 &5.8 &49.9 &5.8 &34.4 \\
EvRT-DETR-B \cite{evrt}& ConvLSTM &52.7 &7.9 &50.1 &7.9 &57.1 \\
Ev-DTAD-B  \cite{ev-dtad} & ConvLSTM &\underline{53.5} &8.4 &50.6 &8.8 &57.8 \\
\cmidrule(l){1-7}
\oursT      & KDA     &52.4 &5.1 &\underline{50.8} &5.5 &22.9 \\
\oursB      & KDA     &\textbf{53.9} &8.4 &\textbf{52.9} &9.6 &56.7 \\
\bottomrule
\end{tabular}
\caption{\textbf{\modelB achieves the best mAP on Gen1 and 1Mpx.}
We report test-set mAP, with baseline numbers from the original
publications. Best in bold, second best underlined. Parameter counts include frozen weights.
Runtimes are our measurements with the released code and weights on an NVIDIA H100, reported as the median of three runs, using compiled mode and autocast to BF16.$^\dagger$~Float32.
$^\ddagger$~Not compiled. Bracketed runtimes are published values as T4 or T4-equivalent times.}
\label{tab:main}
\end{table*}

In this section, we examine what a recurrent memory needs to perform in
event-based detection. All experiments follow the setup of
\cref{sec:framework}, a frozen RT-DETR \cite{detr} with the memory
trained over a long gradient horizon. We present the evidence in the order of \cref{sec:method}, and support each design choice with experiments. 
We start with the detector as a whole (\cref{tab:main,tab:etram}), then isolate the memory mechanism
(\cref{tab:ablation}) and its depth (\cref{tab:depth}), and close with
a synthetic shell game that tests state tracking in isolation
(\cref{fig:shell_game}).
We build two variants using the KDA layer with signed decay \cite{kda,CKDA}, which differ in the frozen
backbone and in the blocks per scale $(L_3, L_4, L_5)$.

\begin{itemize}
  \item \modelT: PResNet-18, $(L_3, L_4, L_5) = (1, 1, 1)$
  \item \modelB: PResNet-50, $(L_3, L_4, L_5) = (4, 9, 2)$
\end{itemize}
All other experiments use the PResNet-50 backbone. \modelT appears only
in the comparison in Tab. \ref{tab:main}.

\subsection{Experiment Setups}
\label{sec:setup}

\paragraph{Datasets and metrics}
Gen1 \cite{Gen1} is recorded from a car at $304 \times 240$ and labels
cars and pedestrians. 1Mpx \cite{1mpx} has more than 14 hours at $1280 \times 720$ with three
classes. eTraM \cite{etram} is recorded by a static camera at the same resolution, and is evaluated on its three categories, vehicles,
pedestrians and micro-mobility. Following EvRT-DETR \cite{evrt}, we halve the resolution
of both to $640 \times 360$ with a bilinear resize. We report COCO mAP \cite{coco} with the Prophesee evaluation protocol \cite{rvt}. Ablations are reported on the validation set after the last training
step. For the comparison with prior work, we select the checkpoint with
the best validation mAP and report it on the test set. At inference, the detector runs over each recording from its first
frame to its last, with the state carried throughout.

\paragraph{Implementation details}
The detector is initialized from an RT-DETR \cite{detr} trained on
single frames and kept frozen. We train the recurrent memory for 20k
steps on Gen1 and 1Mpx with Adam \cite{adam} and a one-cycle schedule
\cite{onecycle}, with a linear warm-up over the first 5\% of the steps
to a peak learning rate of $2 \times 10^{-3}$, and no weight decay. On
eTraM, we train for 8k steps with the same schedule and a lower peak learning rate of $7 \times 10^{-4}$. Each step holds four clips, one per GPU.
Augmentations and losses follow EvRT-DETR \cite{evrt}. We train on four
NVIDIA H100 GPUs.

\subsection{Results}
\paragraph{Gen1 and 1Mpx}
\Cref{tab:main} compares our detectors on Gen1 \cite{Gen1} and 1Mpx
\cite{1mpx} with non-recurrent detectors \cite{ergo,detr} and
with recurrent ones, whose memory sits in the backbone (RED
\cite{1mpx}, ASTMNet
\cite{astmnet}, RVT \cite{rvt}, GET \cite{get}, S5-ViT \cite{SSM-ViT}, SMamba \cite{smamba}, SATE \cite{sate}) or on a frozen detector (EvRT-DETR \cite{evrt}, Ev-DTAD \cite{ev-dtad}). Our 
detector outperforms all of them. The comparison
with EvRT-DETR is the most direct, since both use the same frozen
RT-DETR \cite{detr}. The margin between the two base models on both datasets therefore comes from the temporal module alone.

Another highlight from the results is \modelT. On Gen1\cite{Gen1}, it
matches EvRT-DETR-T \cite{evrt} with a temporal module about five times smaller
than its $3 \times 3$ ConvLSTM \cite{convlstm}. On 1Mpx \cite{1mpx}, it
leads EvRT-DETR-T by 0.9 mAP, and also surpasses
the larger EvRT-DETR-B. The capacity of the memory is thus not what limits a recurrent
detector. We further examine what it does in \cref{sec:ablation} and
\cref{sec:depth}.

\paragraph{eTraM}
On this static-camera benchmark~\cite{etram}, our detector reaches 34.7 mAP (\cref{tab:etram}), ahead of SATE~\cite{sate} (33.0) and SMamba~\cite{smamba} (32.6) and 0.9 points behind Ev-DTAD~\cite{ev-dtad} (35.6). Ev-DTAD introduces a new event representation and a hypergraph spatial module. We hold both components fixed in our study and discuss the gap in \cref{sec:discussions}.

\begin{table}[t]
\centering
\setlength{\tabcolsep}{6pt}
\begin{tabular}{lcc}
\toprule
Model & mAP (\%) & Params (M) \\
\midrule
RVT-B \cite{rvt}          & 29.5 & 18.5 \\
S5-ViT-B \cite{SSM-ViT}   & 29.3 & 17.5 \\
SAST-CB \cite{SAST}       & 30.0 & 18.9 \\
SMamba \cite{smamba}      & 32.6 & 16.7 \\
SATE \cite{sate}          & 33.0 & 26.4 \\
Ev-DTAD-B \cite{ev-dtad}   & \textbf{35.6} & 57.8 \\
\midrule
\oursB                    & \underline{34.7} & 56.7 \\
\bottomrule
\end{tabular}
\caption{\textbf{Comparison on the eTraM test set.} Baseline numbers are as reported in literature. Best in bold, second best underlined.}
\label{tab:etram}
\end{table}

\subsection{More Expressive Updates Improve Detection}
\label{sec:ablation}

\begin{figure*}[t]
  \centering
  \makebox[\linewidth][c]{\hspace*{-1cm}\resizebox{1.03\linewidth}{!}{\input{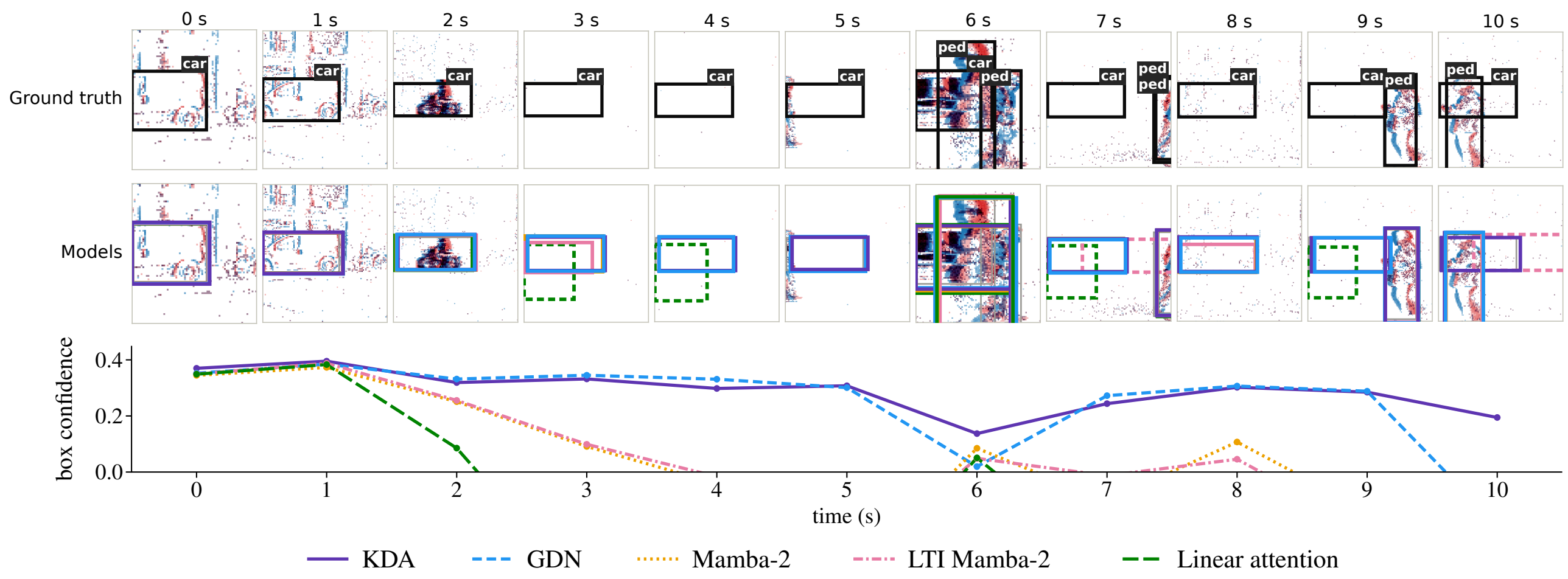}}}
  \caption{\textbf{The delta rule keeps a detection through event
silence.} The car produces no events from 3 to 5\,s. KDA and GDN keep
its box, while Mamba-2, LTI Mamba-2 and linear attention lose it. Top:
ground-truth boxes, one frame per second. Middle: boxes predicted with
each update for the same object. Bottom: confidence of that box above
the detection threshold. A single sequence, shown for illustration.}
  \label{fig:example}
\end{figure*}

One of the main findings of our work is that the accuracy of the
detector follows what the theory of linear recurrences expects in forms of expressivity. The differences between recurrent memories widen on objects that trigger few
events. \Cref{tab:ablation} takes the steps of
\cref{sec:Uniformized_backbone} on Gen1 \cite{Gen1}, with the frozen
RT-DETR \cite{detr} on a PResNet-50 backbone \cite{resnet} and one block per scale. The blocks around the recurrent update and the training are held fixed. We provide an illustration for visual understanding (\cref{fig:example}).

\begin{table}[t]
\centering
\newcommand{\y}{\checkmark}
\begin{tabular}{l c@{\hspace{3pt}}c@{\hspace{3pt}}c cc}
\toprule
& \multicolumn{3}{c}{Recurrence Property}
& All & Faint \\
\cmidrule(lr){2-4} \cmidrule(lr){5-5} \cmidrule(lr){6-6}
Mixer & Decay & Select. & Delta & mAP & AP$_w$ \\
\midrule
No Memory              & & & & 47.9 & 21.5 \\
\midrule
Linear Attn &    &    &    & 53.8 & 46.0 \\
LTI Mamba-2  & \y &    &    & 54.8 & 47.8 \\
Mamba-2      & \y & \y &    & 55.0 & 49.6 \\
GDN         & \y & \y & \y & \underline{55.1} & \textbf{52.3} \\
KDA        & \y & \y & \y & \textbf{55.2} & \underline{52.2} \\
\bottomrule
\end{tabular}
\caption{\textbf{Decay, selectivity and the delta rule each improve
detection, most for faint objects.} Memory updates on Gen1 validation,
one block per scale. Each row adds one property to the row above. Faint
denotes objects in the lowest quarter of event occupancy for their size. AP$_w$ weights each class by its box count, as faint objects are class-imbalanced. Best in bold, second best underlined.}
\label{tab:ablation}
\end{table}

\paragraph{Forgetting}
\Cref{fig:example} follows one object that stops producing events after
2\,s, through two silent gaps separated by other objects crossing its box at 6\,s. We start from the simplest recurrence, linear attention
\cite{linear_attention}, which adds every write to its state without performing any update. It is the first to lose
the box, as soon as the events stop. A constant \cite{s5,retnet} and a
selective decay \cite{mamba2} keep it for about two seconds longer. In \cref{tab:ablation}, linear attention recovers most of what the frame detector
misses, lifting faint AP from 21.5 to 46.0, and both decays improve on it.

\paragraph{Overwriting}
In contrast to the three recurrences without delta rules, layers with overwriting properties
\cite{deltanet,gdn,kda} perform noticeably better in this scene. In
\cref{fig:example}, the box survives the first gap and the crossing at
6\,s, where a large pedestrian disturbs most pixels of the scene. The same holds in general (\cref{tab:ablation}), especially on faint
objects. Extending the decay to a signed, per-channel form
\cite{kda,CKDA} gives the only variant that keeps the box for the full
ten seconds. We discuss this further with the shell game
(\cref{sec:shell_game}).

\subsection{Depth analysis}
\label{sec:depth}

\begin{table}[ht]
\centering
\begin{tabular}{lcc}
\toprule
Blocks & mAP & Params (M) \\
\midrule
No memory & 47.9 & 42.8 \\
\midrule
$(1,1,1)$ & 55.2 & 45.6 \\
\midrule
$(3,3,3)$ & 55.6 & 51.1 \\
$(5,5,5)$ & 55.9 & 56.7 \\
$(5,8,2)$ & \underline{56.1} & 56.7 \\
$(4,9,2)$ & \textbf{56.1} & 56.7 \\
\midrule
$(1,1,1)$, $3\times3$ & 55.2 & 51.1 \\
$(1,1,1)$, $5\times5$ & 55.2 & 62.2 \\
\bottomrule
\end{tabular}
\caption{\textbf{Scaling depth improves detection, whereas expanding
spatial kernels does not.} Validation mAP on Gen1,
with blocks per scale $(L_3, L_4, L_5)$. The last two rows enlarge the kernel of the memory from $1\times1$ to $3\times3$ and $5\times5$.}
\label{tab:depth}
\end{table}
\paragraph{Configurations}
Writing $(L_3,L_4,L_5)$ for the block counts at strides 8, 16, and 32, we compare one block per scale with uniform stacks $(3,3,3)$ and $(5,5,5)$ and two 15-block stacks, $(5,8,2)$ and $(4,9,2)$, that concentrate depth at stride 16, as in hierarchical vision backbones~\cite{resnet,maxvit}. We compare depth scaling with $3\times3$ and $5\times5$ spatial kernels~\cite{convssm,evrt}. 
 
\paragraph{Training deep stacks}
With the learning rate of a single block, deep stacks training falls below their initialization as the learning rate reaches its peak. We attribute this instability to shifts in feature magnitude. The frozen decoder was trained to consume encoder features at their original scale, and the residual stream passes directly to the decoder without normalization. Unlike in language models \cite{gpt2,gdn}, the growth in each branch reaches the decoder directly. To avoid this growth accumulating with deeper networks, we 
follow the Depth-$\mu$P \cite{depthmu} recipe and scale the output of each branch
by $1/\sqrt{L}$, adjusting the peak learning rate of each block
by $\eta/\sqrt{L}$, where $L$ is the number of blocks at that scale.

\paragraph{Results}
Depth scales the temporal model more efficiently than expanding the spatial kernel
(\cref{tab:depth}). Going from one block per scale to $(4, 9, 2)$, our
best variant, improves accuracy by 0.9 mAP, whereas a
$3 \times 3$ or a $5 \times 5$ kernel leaves it unchanged despite increasing parameter count by a similar amount. We observe this more directly in
\cref{sec:shell_game}, where depth extends tracking to more swaps than seen during training.

\subsection{Event-Based Parity Shell Game}
\label{sec:shell_game}

\begin{figure}[t]
\vspace{10pt}
 \centering
\resizebox{\linewidth}{!}{\input{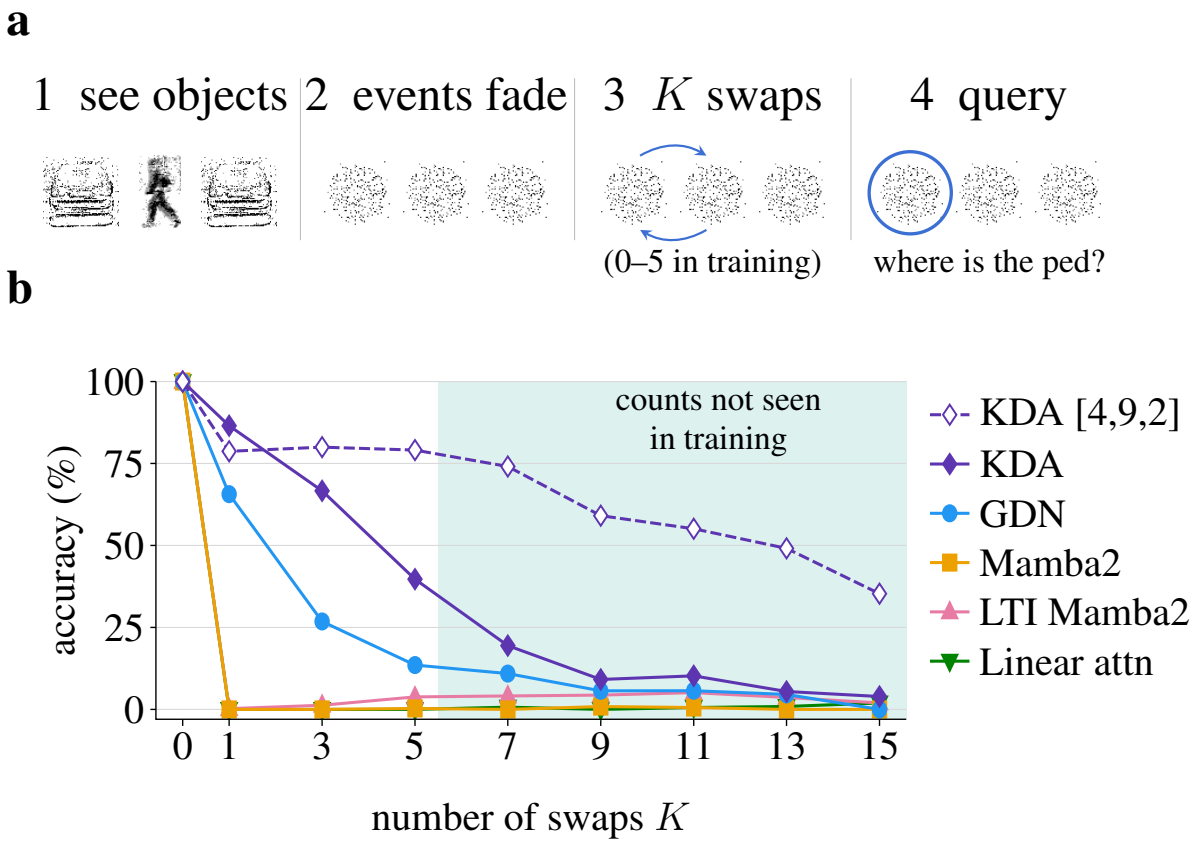}}
\caption{\textbf{Only the delta-rule models follow the swaps.} \textbf{a}, Parity game. One pair of identical discs is
swapped $K$ times, and the answer depends only on the parity of $K$.
\textbf{b}, Accuracy at $K = 0$ and odd $K$. Green marks values of $K$ unseen in
training.}
\label{fig:shell_game}
\end{figure}

In \cref{tab:ablation}, the delta rule improves the detection
of faint objects, whereas overall accuracy changes less clearly. Theory attributes one capability to transitions which involve negative and complex eigenvalues that the other updates
lack, namely state tracking, with permutation composition as the
standard test \cite{illusion_ssm,negative_eigen,CKDA}. To test whether
state tracking is present in our recurrent memory, we construct an event-based parity shell game, a
synthetic task that requires state tracking \cite{cohen2026metbenchmultimodalentitytracking,shin2026video}. In a
shell game, an object is hidden under one of several identical shells,
whose positions are swapped, so that its class can be recovered only by
following every swap from memory.

\paragraph{Task}
Each episode begins with three Gen1 event crops arranged in a row, two
from one class and one from the other~\cite{Gen1}. The
object whose class appears only once is the prediction target. The events fade,
leaving three visually identical discs. The target disc swaps positions
with the same partner $K$ times, while the third disc stays in place. 

At the end of each swap, we select the disc with the highest score for the target class. The prediction is correct if it matches the target's final position. Since each swap moves the target between two positions, a baseline that always selects its starting position is correct exactly when $K$ is even. We report accuracy at $K=0$ to measure localization
before any swaps and at odd $K$ to test whether the model follows the
target away from its starting position (\cref{fig:shell_game}). We
train on zero to five swaps and evaluate on up to fifteen.
\newpage
\paragraph{Results}
All models locate and predict the target with perfect accuracy after the events
fade and before any swaps. Linear attention, LTI Mamba-2, and Mamba-2 \cite{linear_attention,retnet,mamba2}
continue to select the target's starting position, and are therefore correct only when $K$ is even. Only GDN and KDA
follow the target as it swaps position, and KDA detects more
accurately at every nonzero swap count. For both, accuracy declines
as swaps accumulate but remains above the baseline within the
training range. Increasing the depth of
signed KDA from $(1,1,1)$ to $(4,9,2)$ sustains accuracy over more
swaps, including beyond the five seen in training.

\paragraph{Outlook}
Events along moving contours reveal local object shape, a cue long
used in object detection~\cite{hog,dpm}. One hypothesis that state tracking benefits event-based detection could be that each location's
recurrent state retains these cues as events fade and that the delta
rule, through its negative eigenvalues, restores them after a disturbance. 

The parity shell game gives us a controlled setting in which the advantage
of the delta rule widens and its variants, similar in detection,
separate. The second question also remains open in language
modeling, where these recurrences have mainly been studied. We hope the shell game proves useful for studying memory in event-based
detection, and inspires similar probes into when state tracking helps models in
language, vision and other domains.

\section{Discussion}
\label{sec:discussions}
We studied the role of recurrent memory in event-based vision by training linear recurrent modules on frozen spatial features and backpropagating through time on long recordings. The results suggest how a detector updates its memory matters
most when current event evidence is weak. Even a simple recurrence
substantially improves detection of low-event objects, and richer
updates recover more of them. On Gen1, stacking recurrent blocks
improves detection, with little gain from enlarging their spatial
kernels. The resulting \modelB achieves the highest mAP among the
methods compared on Gen1~\cite{Gen1} and 1Mpx~\cite{1mpx}.

Recent work ~\cite{ev-dtad} reports advances in
both event representation and spatial feature extraction, reporting leading
results on eTraM \cite{etram}. As these target orthogonal improvements to components we hold fixed, we expect combining these directions could provide complementary effects. 

More broadly, the recurrences we investigate were mainly studied in language
modeling, and the questions we ask of them come from vision. Continuous
event streams let us measure, object by object, how long event evidence
stays scarce. In turn, the theory of linear recurrence gives us a
principled basis for choosing which memory updates to test. We hope
this connection informs the design of visual memory and motivates
analogous probes of when state tracking helps in other domains.

% \clearpage
% \input{sec/6_LLM}
\section*{Acknowledgments}
Antonio Orvieto is supported by the Hector Foundation. The authors acknowledge the Max Planck Institute for Intelligent Systems for providing the computational resources used in this work. Finally, the authors thank Nadezhda Ilieva for the insights on the uniformized backbone.

{
    \small
    \bibliographystyle{ieeenat_fullname}
    \bibliography{main}
}

\end{document}